\documentclass[10pt,twocolumn,letterpaper]{article}

\usepackage[pagenumbers]{wacv} 

\definecolor{wacvblue}{rgb}{0.21,0.49,0.74}
\usepackage[pagebackref,breaklinks,colorlinks,allcolors=wacvblue]{hyperref}
\usepackage{graphicx,verbatim}
\usepackage{amsmath}
\usepackage{amssymb}
\usepackage{multirow}
\usepackage{booktabs}
\usepackage{xcolor,colortbl}
\usepackage{tocloft}
\usepackage{pifont}
\usepackage{adjustbox}
\usepackage{makecell}
\usepackage[accsupp]{axessibility}

\def\wacvPaperID{24} 
\def\confName{WACV}
\def\confYear{2026}

\title{Gene-DML: Dual-Pathway Multi-Level Discrimination for Gene Expression Prediction from Histopathology Images}

\author{
Yaxuan Song\textsuperscript{1},\enspace 
Jianan Fan\textsuperscript{1}\textsuperscript{\raisebox{-0.5ex}{*}},\enspace
Hang Chang\textsuperscript{2},\enspace
Weidong Cai\textsuperscript{1}\thanks{Corresponding authors.} \\
\textsuperscript{1}The University of Sydney, Australia \qquad \textsuperscript{2}Lawrence Berkeley National Laboratory, USA
\\
{\tt\small yson2999@sydney.edu.au, jianan.fan@sydney.edu.au, hchang@lbl.gov, tom.cai@sydney.edu.au} \\
}

\begin{document}
\maketitle
\begin{abstract}
Accurately predicting gene expression from histopathology images offers a scalable and non-invasive approach to molecular profiling, with significant implications for precision medicine and computational pathology. 
However, existing methods often underutilize the cross-modal representation alignment between histopathology images and gene expression profiles across multiple representational levels, thereby limiting their prediction performance.
To address this, we propose \textbf{Gene-DML}, a unified framework that structures latent space through \textbf{D}ual-pathway \textbf{M}ulti-\textbf{L}evel discrimination to enhance correspondence between morphological and transcriptional modalities. 
The \textbf{multi-scale instance-level} discrimination pathway aligns hierarchical histopathology representations extracted at local, neighbor, and global levels with gene expression profiles, capturing scale-aware morphological-transcriptional relationships. 
In parallel, the \textbf{cross-level instance-group} discrimination pathway enforces structural consistency between individual (image/gene) instances and modality-crossed (gene/image, respectively) groups, strengthening the alignment across modalities.
By jointly modeling fine-grained and structural-level discrimination, Gene-DML is able to learn robust cross-modal representations, enhancing both predictive accuracy and generalization across diverse biological contexts. 
Extensive experiments on public spatial transcriptomics datasets demonstrate that Gene-DML achieves state-of-the-art performance in gene expression prediction. 
The code and processed datasets are available at \href{https://github.com/YXSong000/Gene-DML}{https://github.com/YXSong000/Gene-DML}.
\end{abstract}

\section{Introduction}
\label{sec:intro}

Spatial transcriptomics (ST) has emerged as a transformative technology that enables high-resolution mapping of quantification of mRNA expression (gene expression) within intact tissue architectures~\cite{stahl_2021,anjali_2021,Wang_2024}. 
Unlike bulk RNA sequencing~\cite{wang2025mirror}, single-cell RNA sequencing (scRNA-seq) or single-nucleus RNA sequencing (snRNA-seq), which lacks spatial localization, spatial transcriptomics preserves the anatomical and microenvironmental context of gene expressions, making it invaluable for providing unprecedented insights into the spatial organization of molecular processes~\cite{xie_2024}.
Intuitively, ST technology can be seen as converging on a gene expression matrix that captures the transcriptome at every spot (that is, a pixel, a cell, or a group of cells)~\cite{anjali_2021}, which is divided from large-scale whole-slide images (WSIs).
Traditional ST methods, such as Visium~\cite{stahl_2021}, MERFISH~\cite{chen2015spatially}, seqFISH+~\cite{Eng_2019}, STARmap~\cite{wang2018three} \etal~\cite{ganguly2025merge}, tend to be time-consuming and expensive, relying on specialized instruments and extensive domain knowledge. 
This raises a critical question: how can we leverage recent advances in deep learning to accurately infer spatial gene expression directly from WSIs?

Several recent studies have made substantial progress in gene expression prediction from WSIs~\cite{Mejia_2024,yang_2024}.
Unimodal approaches~\cite{he_2020,pang_2021,zeng_2022,Yang_2023,chung_2024} typically employ supervised learning to predict gene profiles from WSI tiles, focusing on improving feature extraction to learn strong visual representations.
However, despite the use of powerful image encoders, these unimodal methods often fail to incorporate gene-specific morphological information.
The absence of alignment with gene expression weakens the model’s ability to capture meaningful biological signals, often resulting in overfitting to high-dimensional gene expression counts and weak generalization across patient samples.

To explore image-gene relationships more effectively, bimodal methods~\cite{min_2024,huang_2024,lin_2024} co-embed WSI tiles and gene expression profiles in a shared space.
While these approaches enable cross-modal learning, they typically model only single-level, pairwise relationships between individual tiles and gene instances.
By treating each image-gene pair independently, they fail to capture the hierarchical alignment inherent in neither WSI tile nor gene expression profile instances, triggering overfitting to the instance-level noise.
Thus, multi-level alignments between image and gene modalities are essential for constructing a robust and expressive latent space that faithfully encodes both histopathology morphology~\cite{10.1007/978-3-031-73464-9_17,8363568,9731234} and gene expression; otherwise, the model's generalization and the learning of biologically coherent representations are compromised.

To this end, we propose \textbf{Gene-DML}, a novel framework that enhances \textbf{Gene}-image correspondence by performing \textbf{D}ual-pathway \textbf{M}ulti-\textbf{L}evel discrimination.
It jointly captures the scale-aware morphological-transcriptional relationship between multi-scaled WSI tiles and gene expression profiles at the instance level and instance-group level, that is, WSI tile instances aligned with clustered gene-expression groups and vice versa. 
By jointly modeling fine-grained and high-level discrimination, Gene-DML enables the learning of robust cross-modal representations, enhancing both predictive accuracy and generalization across diverse biological contexts. Contributions are summarized as:

\begin{itemize}
    \item We propose Gene-DML, a novel framework that models dual-pathway multi-level discrimination to enhance spatial and morphological alignment between histopathology images and gene expression profiles.
    \item Gene-DML integrates multi-scale instance-level discrimination pathway to align gene expression with WSI features extracted at local, neighbor, and global scales, enabling scale-aware cross-modal representation learning.
    \item In parallel, Gene-DML incorporates the cross-level instance-group discrimination pathway, which jointly enforces consistency of instance-level and group-level across image and gene modalities, improving semantic structure in the shared latent space.
    \item Through comprehensive experiments on public ST datasets and additional external tests on Visium data, Gene-DML outperforms state-of-the-art methods on multiple datasets, achieving superior prediction accuracy and generalization across heterogeneous patient samples.
\end{itemize}

\section{Related Work}
\label{sec:related}

\subsection{Gene Expression Prediction}

Early methods for spatial gene expression prediction from histology~\cite{fan2024learning}, such as ST-Net~\cite{he_2020}, employed a transfer learning strategy by fine-tuning a DenseNet121 model~\cite{huang_2017} pre-trained on ImageNet to capture image features and map WSI tiles to gene expression labels. 
Building upon this, HisToGene~\cite{pang_2021} integrates Vision Transformers (ViT) to capture global contextual dependencies between image tiles within a WSI.
Further advancing this direction, Hist2ST~\cite{zeng_2022} incorporates ConvMixer~\cite{Trockman_2022} and a graph convolution network (GCN) post-initially for enhanced image tile embeddings by aggregating neighborhood information.
Recent multi-scale models, TRIPLEX~\cite{chung_2024} and M2OST~\cite{wang2025m2ost} improved performance by integrating spot-, neighborhood-, and slide-level features. These architectures leveraged hierarchical modeling and fusion techniques to capture both fine-grained morphology and global tissue context.
EGN~\cite{Yang_2023} is an alternative paradigm in which image embedding is optimized by selecting the most analogous exemplars from the target spot within a WSI to enhance gene expression prediction. 
Despite these advances, these unimodal methods lack semantic alignment with gene expression and struggle to generalize across samples.
Bimodal contrastive learning has gained traction. To explore the image-gene relationship for more scalable and broader potential approaches, bimodal models, BLEEP~\cite{xie_2024} and MclSTExp~\cite{min_2024} with Transformer encoders align image and gene embeddings through contrastive loss, enabling gene querying via latent similarity. 
RankByGene~\cite{huang_2024} and ST-Align~\cite{lin_2024} further improved alignment through ranking consistency and large-scale pretraining, while Stem~\cite{zhu2025diffusion} explored generative modeling using diffusion techniques~\cite{wan2025tedvit}.
Existing contrastive or generative methods focus mainly on pairwise alignment and overlook deeper morphological-transcriptional correspondences across image scales, which is essential for biological semantic understanding.

\subsection{Multi-scaled Image Feature Extraction}

Several recent works have explored multi-scale modeling for WSIs to capture both fine-grained and contextual features~\cite{fang2025scsamdebiasingmorphologydistributional}. HIPT by Chen \etal~\cite{chen2022scaling} employs hierarchical self-supervised learning to learn patch-level and region-level features via contrastive objectives, enabling effective representation learning across image resolutions. 
TRIPLEX~\cite{chung_2024} is proposed by integrating multi-resolution features from target, neighbor, and global tissue levels using a supervised fusion strategy.
M2OST~\cite{wang2025m2ost} leverages multi-scale WSI patches with Transformer-based many-to-one regression for accurate gene prediction. 
Cross-scale MIL architectures have also shown promise. 
MEGT~\cite{ding2023multi} and DSF-WSI~\cite{wang2023dual} adopt dual-branch architectures with separate high- and low-resolution encoders, fusing outputs using graph‑Transformer modules or masked‐jigsaw tasks.
Despite these advances, existing methods suffer from the following limitations: (1) most focus exclusively on unimodal representation learning without leveraging alignment of transcriptomic modality data, and (2) fusion-based designs often lack contrastive objectives that explicitly enforce scale-consistent representation. In contrast, Gene-DML incorporates multi-scale instance-level discrimination that explicitly aligns local, neighbor, and global image features with gene expression, enriching cross-modal representations and enhancing semantic consistency across scales.

\begin{figure*}[t]
\centering
\includegraphics[width=\textwidth]{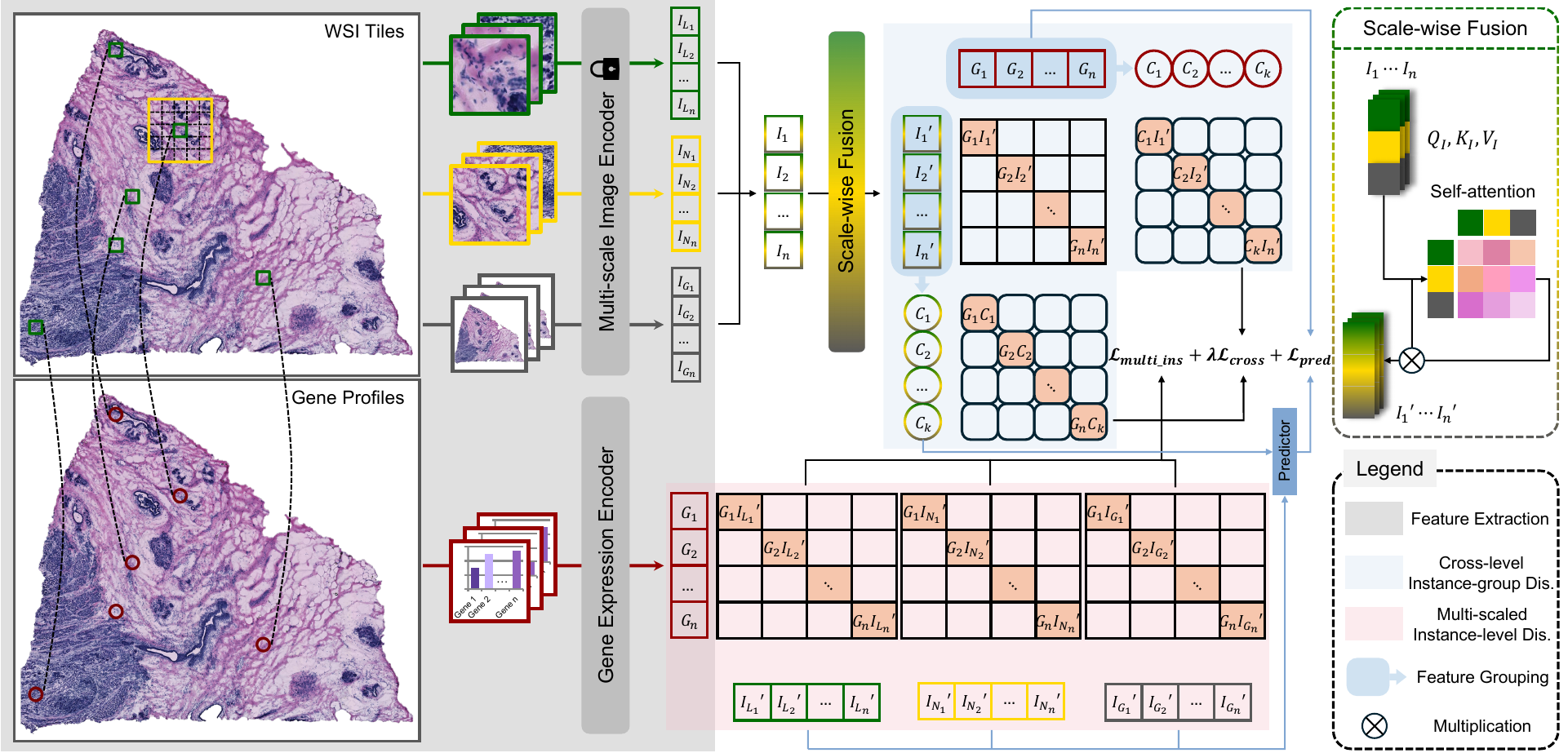}
\caption{An overview of Gene-DML framework. Pairs of WSI tiles and gene profiles are aligned from both the \textbf{multi-scale instance-level} and the \textbf{cross-level instance-group} discrimination. 
An illustration of feature grouping with instance-group alignment is shown in~\cref{fig2}.} 
\label{fig1}
\vspace{-0.4cm}
\end{figure*}

\subsection{Cross-Level Instance-Group Discrimination}

Cross-Level Discrimination (CLD)~\cite{wang_2021} is proposed to improve instance-level contrastive learning by incorporating dynamic group-level structure within a single modality. 
By pairing each instance with its group centroid, CLD mitigates over-separation of semantically similar samples and improves representation stability. 
Subsequent work has adapted instance-group techniques within a single modality and for video-text alignment, \eg, CrossCLR~\cite{zolfaghari2021crossclr}, TCL~\cite{li2022twin}, by incorporating grouping or de-biasing false negatives, these methods empirically show improved retrieval, classification, and stability~\cite{li2022twin}. However, they remain unimodal or restricted to intra-modality alignment.
In medical imaging and genomics, multimodal frameworks like ContIG~\cite{taleb2022contig} and MGI~\cite{li2022twin} have adopted instance-level contrastive learning across modalities (\eg, image-genetic or image-omics pairs), but these lack mechanisms to model group-level relationships or latent clustering in cross-modal~\cite{chen2024gaussian}.
They typically use pairwise contrastive objectives akin to CLIP~\cite{radford2021learning,wang2023mosaic}, which may overemphasize instance-level uniqueness and discount shared biological patterns.
To bridge these gaps, Gene-DML integrates a cross-level instance-group discrimination mechanism within two modalities: images and transcriptomics. 
It aligns each image instance with gene-expression groups and vice versa, ensuring the model learns shared semantic structure between morphology and transcriptional activity. 
This bidirectional group alignment improves robustness to noise, enhances positive/negative contrastive signal, and supports generalization to unseen tissue domains.
\section{Methodology}

In this study, we explore how to effectively leverage the alignment information between WSIs and gene expression profiles via a dual-pathway discrimination strategy to enhance gene expression prediction.
Specifically, our objective is to train a bimodal contrastive learning model that captures semantic alignment information between morphological features and gene functions, which is distilled using WSI tiles, represented as $\mathbf{I} \in \mathbb{R}^{N \times H \times W \times 3}$, to predict the gene expression levels for each spot, represented as $\mathbf{G} \in \mathbb{R}^{N \times M}$, in a $N$-spot WSI.
Here, $M$ denotes the number of gene types whose expression levels are to be predicted, and $H$ and $W$ signify the height and width of each spot image, respectively.
An overview of the framework we proposed, Gene-DML, is illustrated in~\cref{fig1}.

\subsection{Multi-scale Instance-level Alignment}

\label{multi-scale}

\subsubsection{Multi-scale Image Instance Feature}
A single spot’s gene expression is influenced not only by its WSI tile structure but also by its surrounding environment~\cite{chung_2024}. 
Each WSI is split into tiles centered on spatial transcriptomics spots, extracted at three spatial scales: local tiles $I_L$ focus on local-level detail, neighbor tiles $I_N$ capture microenvironment context, and the global scale $I_G$ represents the WSI's global tissue structure. These tiles are encoded via a shared Multi-Scale Image Encoder.
Specifically, we utilize the UNI~\cite{chen2024uni} model, pretrained using more than 100 million H\&E-stained WSIs, for histopathology image feature extraction.
In order to avoid the overfitting problem, the base feature extraction model UNI is kept frozen to ensure the generalization of Gene-DML. 
The features from all three scales are individually projected into $512$ dimensions via separate linear layers.

\vspace{-1em}

\paragraph{Local-scale.}
For local-scale (green square in~\cref{fig1}) feature extraction, the pretrained UNI model is utilized to process $N$ tiles with a size of $224\times224$ for each WSI. These features are extracted from the forward feature of the UNI model. These features are embedded into $265$ features with $1024$ dimensions for each.

\vspace{-1em}

\paragraph{Neighbor-scale.}
As for the neighbor-scale (yellow square in~\cref{fig1}), to preserve the structural integrity of the tissue within the WSI, we utilize a region of size $1120\times1120$ pixel tiles of centered spot rather than the surrounding $5$ tiles. These neighbor-scale features are embedded into $25$ $1024\times1024$ features by using the UNI model.
Since the UNI model is frozen, a series of self-attention blocks are applied on neighbor-scale features, which are kept updated for better model learning.  

\vspace{-1em}

\paragraph{Global-scale.}
To efficiently capture global spatial context in WSIs without incurring high computational costs, inspired by~\cite{chung_2024}, the global-scale representation is concatenated with all local-scale tile features extracted by the pre-trained UNI model, which is then encoded through the transformer block for the ultimate global-scale token.

\vspace{-1em}

\paragraph{Scale-wise Fusion.}
To effectively integrate morphological information across scales, multi-scale image features captured at local, neighbor, and global scales are concatenated for each spot along the channel dimension $I_n=\{I_{L_n}, I_{N_n}, I_{G_n}\}$ and passed into a Scale-wise Fusion module designed to model cross-scale dependencies.
This module adopts a self-attention mechanism that dynamically reweights information across levels, enabling the network to integrate fine-grained cellular patterns with broader tissue context. The fused outputs $I^{E_\text{ins}}_i=\{I_1’, I_2’, …, I_n’\}\text{, where } i=1,..., n$, retain scale-specific attributes and are subsequently disentangled for downstream alignment with gene expression features. This process ensures semantically enriched, scale-aware representations for each spot, facilitating more robust cross-modal learning.

\subsubsection{Gene Expression Instance Feature}
In parallel, to model transcriptomic information in the shared latent space, we employ a lightweight yet expressive fully connected network as the gene expression encoder. Each spatial gene expression profile is passed through the FCN to produce a dense embedding:
\begin{align}
G^{E_{\text{ins}}}_i = \text{FFN} \circ \text{GeneEncoder}(G_i) \in \mathbb{R}^d,
\end{align}
where $i=1,...,n$. Here, GeneEncoder first maps raw gene profiles into an intermediate embedding space of dimension $d$, capturing high-level transcriptomic structure, consisting of two linear layers with GELU~\cite{hendrycks2016gaussian} activation and dropout for regularization. This is followed by a FeedForward (FFN) network consisting of two linear layers with a non-linear activation and dropout, which refines the representation and projects it into the shared latent space.
This design allows Gene-DML to learn instance-level gene features that align with corresponding histology-derived representations.

\subsection{Cross-Level Instance-Group Alignment}

\begin{figure*}[t]
\centering
\includegraphics[width=0.85\textwidth]{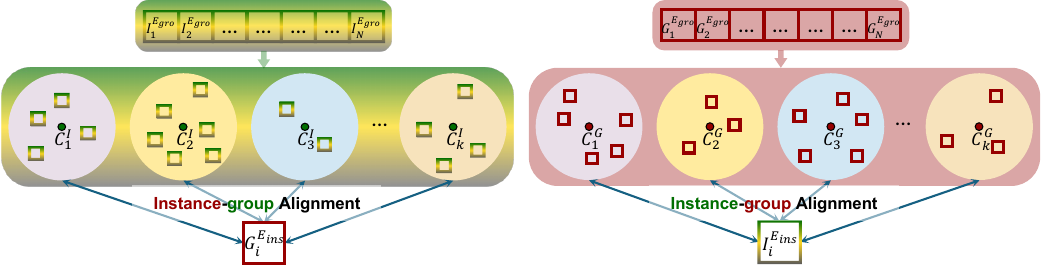}
\caption{Detailed illustration of feature grouping intuition. The image group features (gradient color squares) and gene group features (red squares) are grouped into $k$ groups with centroids $\{C^I_1, \dots C^I_k\}$ and $\{C^G_1, \dots C^G_k\}$.} 
\label{fig2}
\vspace{-0.2cm}
\end{figure*}

\subsubsection{Feature Grouping}
\label{sec:feature_grouping}
To enrich broader semantic alignment information, which enforces the transcriptional activity of a certain tile of WSI close to a group of tiles with similar morphology and vice versa, between the morphological features of WSI tiles and the functional meaning of gene expression, we integrate a \textbf{feature grouping} branch for both histopathology and gene expression modalities.
As illustrated in the light blue section of~\cref{fig1}, for both feature grouping branches of image and gene, group features are derived from the instance-level embeddings $\{I^{E_{ins}}_i\}$ and $\{G^{E_{ins}}_i\}$ (\cref{multi-scale}) via a fully connected transformation:
\begin{equation}
I^{E_{\text{clu}}}_i = \text{FC}(I^{E{_\text{ins}}}_i)\text{ and } G^{E{_\text{clu}}}_i = \text{FC}(G^{E{_\text{ins}}}_i).
\end{equation}
This branch serves as a preparatory step for downstream clustering by projecting instance features into a space optimized for group structure discovery. Compared to the instance embeddings ($I^{E{_\text{ins}}}_i$ and $G^{E{_\text{ins}}}_i$) that may contain modality-specific noise or fine-grained variations, the learned group features emphasize higher-level semantic consistency, making them more suitable for clustering. This facilitates more coherent dual-view contrastive alignment not only at the instance level but also across clusters that capture shared morphological or transcriptional patterns.

To structurally organize the latent space and uncover shared biological semantics, we perform modality-specific feature grouping based on the group features $\{I^{E_{\text{clu}}}_i\}$ and $\{G^{E_{\text{clu}}}_i\}$. 
Specifically, the $K$-means strategy~\cite{Ye_2007} is applied to unit-normalized vectors to form $k$ local clusters in each modality. 
This yields sets of centroids $\{C^I_1, \dots, C^I_k\}$ and $\{C^G_1, \dots, C^G_k\}$, which are $\{C_i\}$ in gradient-colored circles and $\{C_i\}$ in red circles, respectively, as shown in~\cref{fig1}.

\subsubsection{Cross-Level Alignment}
As for cross-level instance-group alignment, each instance-level embedding $I^{E_{\text{ins}}}_i$ (or $G^{E_{\text{ins}}}_i$) is assigned to its nearest gene (or image) cluster centroid via a mapping $\mathcal{A}(i) = j$. As illustrated in~\cref{fig2}, we then compute instance-to-group similarities bidirectionally across modalities:
\begin{itemize}
    \item Aligning each gene instance $G^{E_{\text{ins}}}_i$ with its corresponding visual group centroid $C^I_j$;
    \item Aligning each image instance $I^{E_{\text{ins}}}_i$ with its corresponding gene group centroid $C^G_j$.
\end{itemize}
This instance-group alignment enables Gene-DML to reinforce semantic consistency across heterogeneous domains while mitigating overfitting to noisy or isolated instance-level correlations. 
By leveraging local group structure in both vision and gene spaces, the model gains robustness and improved generalization to out-of-distribution samples.

\subsection{Loss Component}

Our method follows the CLIP~\cite{radford2021learning} framework that maximizes the cosine similarity of the image and gene embeddings of the $N$ real pairs while minimizing the cosine similarity of the embeddings of the $N^2 - N$ incorrect pairings at the image-gene instance level.
At the multi-scale instance-level, the contrastive loss is applied on WSI embeddings from different scales with gene embedding.
At the instance-group level, Gene-DML applies a batch-wise contrastive loss across views between group features $I^{E_\text{ins}}_i$ and group centroids $C^G$, and vice versa for $G^{E_\text{ins}}_i$ (with $C^I$), to supervise the consistency of grouping relationships.

\subsubsection{Multi-scale Instance-level Loss}

Inspired by the contrastive loss used in~\cite{xie_2024}, Gene-DML computes the internal multi-scaled similarities between the image or the gene expression features in advance, denoted as $\mathbf{T}$, increasing the coherence of the latent space for joint embedding.
Then, cross-entropy loss is applied to align the multi-scaled image and gene expression instance features:
\begin{align}
\mathbf{T^s} = \sigma &\left( \frac{
\text{sim}(I^{E_\text{ins}}_{s}, I^{E_\text{ins}}_{s}) + \text{sim}(G^{E_\text{ins}}, G^{E_\text{ins}})
}{2 \cdot \tau} \right), \\
\mathcal{L}_{\text{multi\_ins}} = & \frac{1}{3} \sum_{s} \frac{1}{N} \sum_{i=1}^{N} \bigg[ 
- \sum_{j=1}^{N} \mathbf{T^s}_{ij} \log \sigma \left( \text{sim}(I^{E_{\text{ins}}}_{s_i}, G^{E_{\text{ins}}}_{j}) \right) \nonumber \\
& - \sum_{j=1}^{N} \mathbf{T^s}_{ji} \log \sigma \left( \text{sim}(I^{E_{\text{ins}}}_{s_j}, G^{E_{\text{ins}}}_{i}) \right) \bigg],
\end{align}
where $s$ denotes the scale of WSI tiles $s \in \{L, N, G\}$, $\sigma(\cdot)$ denotes the softmax function, and $\text{sim}(A, B)=AB^T$.

\subsubsection{Instance-group Loss}
\label{sec:i-gloss}
The discrimination between instance and group can be regarded as minimizing the cross-entropy between the hard grouping centroids and the predicted grouping assignment.
Specifically, the cross-entropy between the group centroids $C^G_i$ computed by $G^{E_\text{clu}}_i$ and the soft grouping assignment predicted from $I^{E_\text{ins}}_i$ should be minimized, and vice versa for $G^{E_\text{ins}}_i$.
Thus, we have the instance-group cross-level loss that validates feature grouping across different views:
\begin{align}
\mathcal{L}_{\text{cross}} = 
& \frac{1}{N} \sum_{i=1}^{N} \bigg[ - \sum_{j=1}^{N} \mathbf{T^I}_{ij} \log \sigma \left( \text{sim}(I^{E_{\text{ins}}}_{i}, C^G_{j}) \right) \nonumber \\
& - \sum_{j=1}^{N} \mathbf{T^G}_{ji} \log \sigma \left( \text{sim}(G^{E_{\text{ins}}}_{j}, C^I_{i}) \right) \bigg],
\end{align}
where $\mathbf{T^I} = \frac{
\text{sim}(I^{E_{\text{ins}}}_{i}, C^G_{j})}{\tau_{I\text{-}G}}$, 
$\mathbf{T^G} = \frac{
\text{sim}(G^{E_{\text{ins}}}_{j}, C^I_{i})}{\tau_{I\text{-}G}}$, and $\tau_{I\text{-}G}$ denotes the temperature of instance-group discrimination.

To facilitate downstream inference of gene expression, we append a prediction head atop the fused image representations $\{I_i'\}$. This module maps the learned self-supervised features to predicted gene expression profiles $\hat{G}_i$, and is optimized using a standard mean squared error (MSE) loss:
\begin{equation}
\mathcal{L}_{\text{pred}} = \frac{1}{N} \sum_{i=1}^{N} \left\| \hat{G}_i - G_i \right\|_2^2.
\end{equation}
By combining the instance-level and instance-group losses with a task-specific prediction loss, Gene-DML has the ability to learn the alignments between WSIs and spatial gene expressions in terms of different semantic levels with the weight $\lambda$ of instance-group loss:
\begin{equation}
\mathcal{L_\text{total}} = \mathcal{L_\text{multi\_ins}} + \lambda\mathcal{L_\text{cross}} + \mathcal{L_\text{pred}}.
\end{equation}

\section{Experiments and Results}

\begin{table*}[t!]
\resizebox{\textwidth}{!}{%
\begin{tabular}{c|ccc|ccc|ccc}
\toprule

\multirow{2}[2]{*}{Model}    & \multicolumn{3}{c|}{HER2ST} & \multicolumn{3}{c|}{STNet} & \multicolumn{3}{c}{skinST} \\
\cmidrule(lr){2-4}
\cmidrule(lr){5-7}
\cmidrule(lr){8-10}

& MSE$\downarrow$ & PCC(A)$\uparrow$ & PCC(H)$\uparrow$ & MSE$\downarrow$ & PCC(A)$\uparrow$ & PCC(H)$\uparrow$ & MSE$\downarrow$ & PCC(A)$\uparrow$ & PCC(H)$\uparrow$\\
\midrule
ST-Net \cite{he_2020} 20'& $0.260\pm0.04$ & $0.194\pm0.11$ & $0.345\pm0.16$ &
$0.209\pm0.02$ & $0.116\pm0.06$ & $0.223\pm0.10$ &
$0.294\pm0.07$ & $0.274\pm0.08$ & $0.382\pm0.08$ \\
HistoGene \cite{pang_2021} 21'& $0.314\pm0.09$ & $0.168\pm0.12$ & $0.302\pm0.19$ &
$0.194\pm0.05$ & $0.100\pm0.05$ & $0.219\pm0.12$ &
$0.270\pm0.09$ & $0.133\pm0.06$ & $0.261\pm0.13$ \\
Hist2ST \cite{zeng_2022} 22'& $0.285\pm0.08$ & $0.118\pm0.10$ & $0.248\pm0.17$ &
$\underline{0.181}\pm0.02$ & $0.044\pm0.02$ & $0.099\pm0.03$ &
$1.291\pm0.65$ & $0.004\pm0.01$ & $0.053\pm0.01$ \\
EGN \cite{Yang_2023} 23'& $0.241\pm0.06$ & $0.197\pm0.11$ & $0.328\pm0.17$ &
$0.192\pm0.02$ & $0.111\pm0.05$ & $0.203\pm0.09$ &
$0.281\pm0.08$ & $0.281\pm0.06$ & $0.388\pm0.06$ \\
BLEEP \cite{xie_2024} 23'& $0.277\pm0.05$ & $0.151\pm0.11$ & $0.277\pm0.16$ &
$0.235\pm0.02$ & $0.095\pm0.05$ & $0.193\pm0.10$ &
$0.297\pm0.05$ & $0.269\pm0.07$ & $0.396\pm0.08$ \\
TRIPLEX~\cite{chung_2024} 24'& $\underline{0.228}\pm0.07$ & $\underline{0.314}\pm0.14$ & $\underline{0.497}\pm0.17$ & $0.202\pm0.02$ & $\underline{0.206}\pm0.07$ & $0.352\pm0.10$ &
$\underline{0.268}\pm0.09$ & $\underline{0.374}\pm0.07$ & $\underline{0.490}\pm0.07$ \\
M2OST~\cite{wang2025m2ost} 25'& $0.302\pm0.06$ & $0.231\pm0.10$ & $0.410\pm0.17$ & $0.278\pm0.03$ & $0.201\pm0.07$ & $\underline{0.353}\pm0.10$ &
$0.271\pm0.02$ & $0.300\pm0.05$ & $\underline{0.490}\pm0.05$ \\
\rowcolor{gray!25} Gene-DML (ours) & $\textbf{0.210}\pm0.03$ & $\textbf{0.331}\pm0.07$ & $\textbf{0.541}\pm0.08$ &
$\textbf{0.179}\pm0.02$ & $\textbf{0.237}\pm0.09$ & $\textbf{0.384}\pm0.08$ &
$\textbf{0.237}\pm0.04$ & $\textbf{0.433}\pm0.03$ & $\textbf{0.520}\pm0.07$ \\
\bottomrule
\end{tabular}}
\caption{Comparison of cross-validation experiments with baselines. PCC(A) denotes the average PCC of all $250$-genes, and PCC(H) denotes the average PCC of the top-$50$ (high) predictive genes.
The \textbf{best results} are in bold; the \underline{second-best results} are underlined.}
\label{tab1_baseline}
\vspace{-0.2cm}
\end{table*} 

\subsection{Dataset and Implementation Details}
In this study, we conducted experiments on three ST datasets that are collected in the HEST-1k~\cite{Jaume_2025} benchmark: HER2ST~\cite{Andersson_2021}, STNet~\cite{he_2020}, and skinST~\cite{ji_2020}.
Specifically, two breast cancer ST datasets -- HER2ST, which focuses exclusively on HER2-positive breast tumor subtypes, and STNet, which includes a broader range of breast cancer subtypes -- provide approximately $\mathbf{13,620}$ and $\mathbf{68,050}$ spatially resolved spots, respectively.
Additionally, skinST contains approximately $\mathbf{23,205}$ spots from spatial profiling of skin tissue sections.
Each spot comprises its coordinates on the WSI sample and the corresponding gene expression profiles with around $20,000$ genes for breast and skin cancer.
To prevent data leakage, we ensure that samples from the same patient do not appear in both the training and testing datasets.
Following~\cite{chung_2024}, we adopt the one-patient-one-fold cross-validation (CV) approach for the HER2ST (8-fold) and skinST (4-fold) datasets; for STNet, which comprises $68$ samples from $23$ patients, we perform an 8-fold CV, ensuring that all samples from the same patient are assigned to the same fold. 
To evaluate the generalization performance of Gene-DML, we perform external tests on $3$ breast Visium data and $2$ skin Visium data.
In our experiments, we use Pearson Correlation Coefficient (PCC), Mean Squared Error (MSE), and Mean Absolute Error (MAE) as evaluation metrics:
\begin{align}
\small
&\text{PCC} = \frac{\sum_{i=1}^{N} (y_i - \bar{y})(\hat{y}_i - \bar{\hat{y}})}{\sqrt{\sum_{i=1}^{N} (y_i - \bar{y})^2} \sqrt{\sum_{i=1}^{N} (\hat{y}_i - \bar{\hat{y}})^2}},\\
& \text{MSE} = \frac{1}{N} \sum_{i=1}^{N} (y_i - \hat{y}_i)^2, \text{ and}\\
& \text{MAE} = \frac{1}{N} \sum_{i=1}^{N} |y_i - \hat{y}_i|.
\end{align}
We report both the mean PCC for all genes (PCC(A)) and the mean PCC for highly predictive genes (PCC(H)).
Specifically, PCC is computed for each gene across all spatial spots within a sample, and the highly predictive genes (HPG) are identified as the top-$50$ genes selected based on their average ranking across all CV folds.

At the data-preprocessing stage, we crop patches with $224\times224$ pixel patches centered on each spatial spot from WSIs to construct image tiles.
For gene expression profiles, we select 250 genes per dataset (details refer to the supplementary), applying spot-wise normalization followed by log transformation according to~\cite{he_2020}.
For feature grouping, we utilize K-means~\cite{Ye_2007} method. 
Gene-DML is optimized using the Adam optimizer with an initial learning rate of $1e^{-4}$. The learning rate is decayed by a factor of $0.95$ every $20$ epochs using a StepLR scheduler.
The batch size is set to $256$ during training, and we report the mean of the model's performance evaluated on all spots in every validation WSI during testing.
All the experiments are trained on an NVIDIA RTX A6000 (48GB) GPU.

\subsection{Cross-validation Performance of Gene-DML}

To evaluate the performance of Gene-DML, we conducted cross-validation experiments on the public ST datasets, HER2ST, STNet, and skinST, and compared the spatial gene expression result predicted by Gene-DML with recent-year existing methods as baselines: ST-Net~\cite{he_2020}, EGN~\cite{Yang_2023}, BLEEP~\cite{xie_2024}, HistoGene~\cite{pang_2021}, Hist2ST~\cite{zeng_2022}, TRIPLEX~\cite{chung_2024}, and M2OST~\cite{wang2025m2ost}.
We re-implement the baseline methods by their public code repositories under the same CV fold splitting settings on the same datasets. 

\begin{table*}
\resizebox{\textwidth}{!}{%
\begin{tabular}{c|cccc|cccc|cccc}
\toprule[1pt]
\multirow{2}[2]{*}{Model} & \multicolumn{4}{c|}{Breast Visium-1} & \multicolumn{4}{c|}{Breast Visium-2} & \multicolumn{4}{c}{Breast Visium-3} \\
\cmidrule(lr){2-5}
\cmidrule(lr){6-9}
\cmidrule(lr){10-13}
& MSE$\downarrow$ & MAE$\downarrow$ & PCC(A)$\uparrow$ & PCC(H)$\uparrow$ & MSE$\downarrow$ & MAE$\downarrow$ & PCC(A)$\uparrow$ & PCC(H)$\uparrow$ & MSE$\downarrow$ & MAE$\downarrow$ & PCC(A)$\uparrow$ & PCC(H)$\uparrow$ \\
\midrule
ST-Net \cite{he_2020} 20'& 0.423 & 0.505 & -0.026 & -0.000 &
0.395 & 0.492 & 0.091 & 0.193 &
0.424 & 0.508 & -0.032 & 0.008 \\
EGN \cite{Yang_2023} 23'& 0.421 & 0.512 & 0.003 & 0.024 &
0.328 & 0.443 & 0.102 & 0.157 &
0.303 & 0.425 & 0.106 & 0.220 \\
BLEEP \cite{xie_2024} 23'& \underline{0.367} & \underline{0.470} & 0.106 & 0.221 &
0.289 & \textbf{0.406} & 0.104 & 0.260 &
\underline{0.298} & 0.415 & 0.114 & 0.229 \\
TRIPLEX~\cite{chung_2024} 24'& \textbf{0.351} & \textbf{0.464} & \underline{0.136} & \underline{0.241} &
\textbf{0.282} & 0.407 & \underline{0.155} & \underline{0.356} &
\textbf{0.285} & \underline{0.410} & \underline{0.118} & \underline{0.282} \\
\rowcolor{gray!25} Gene-DML (ours) & 0.410 & 0.512 & \textbf{0.160} & \textbf{0.415} &
\underline{0.289} & \underline{0.407} & \textbf{0.192} & \textbf{0.411} &
0.301 & \textbf{0.408} & \textbf{0.155} & \textbf{0.393} \\
\bottomrule[1pt]
\end{tabular}}
\caption{Comparison of generalization performance with baselines. The \textbf{best results} are in bold; the \underline{second-best results} are underlined.}
\label{tab2_general_main}
\vspace{-0.2cm}
\end{table*} 

In comparison with existing works, as shown in~\cref{tab1_baseline}, Gene-DML consistently achieves the best performance across all metrics, MSE, PCC(A), and PCC(H), demonstrating its superiority in accurately capturing spatial gene expression patterns.
Specifically, for HER2ST dataset, Gene-DML obtains the lowest MSE of $0.210$, outperforming the second-best TRIPLEX ($0.228$) by $0.018$.
In terms of PCC(A), our method achieves $0.331$, which surpasses TRIPLEX ($0.314$) by $+0.017$, and shows the most significant gain in PCC(H) at $0.541$, improving by $+0.044$ over TRIPLEX ($0.497$).
As for STNet dataset, Gene-DML achieves the best MSE of $0.179$, a $0.002$ reduction from Hist2ST ($0.181$), and attains a notable gain in PCC(A) with $0.237$, improving upon TRIPLEX ($0.206$) by $+0.031$. In high-predictive gene correlation (PCC(H)), our method leads $0.384$, ahead $0.353$ of M2OST  by $+0.031$.
On the skin dataset skinST, Gene-DML achieves the most substantial improvements. Its MSE of $0.237$ is better than $0.268$ of TRIPLEX by $+0.031$. In PCC(A), it scores $0.433$, significantly outperforming the second-best TRIPLEX ($0.374$) by $+0.059$. For PCC(H), Gene-DML leads with $0.520$, an improvement of $+0.030$ over M2OST and TRIPLEX ($0.490$).
Due to space limitations, all MAE results are provided in the supplementary material.

Across all datasets and metrics, Gene-DML consistently sets a new performance benchmark, which reflects its capacity to capture biologically critical gene patterns. The consistent outperformance validates the strength of our dual-pathway multi-level alignments.

\begin{figure}
    \centering
    \includegraphics[width=\linewidth]{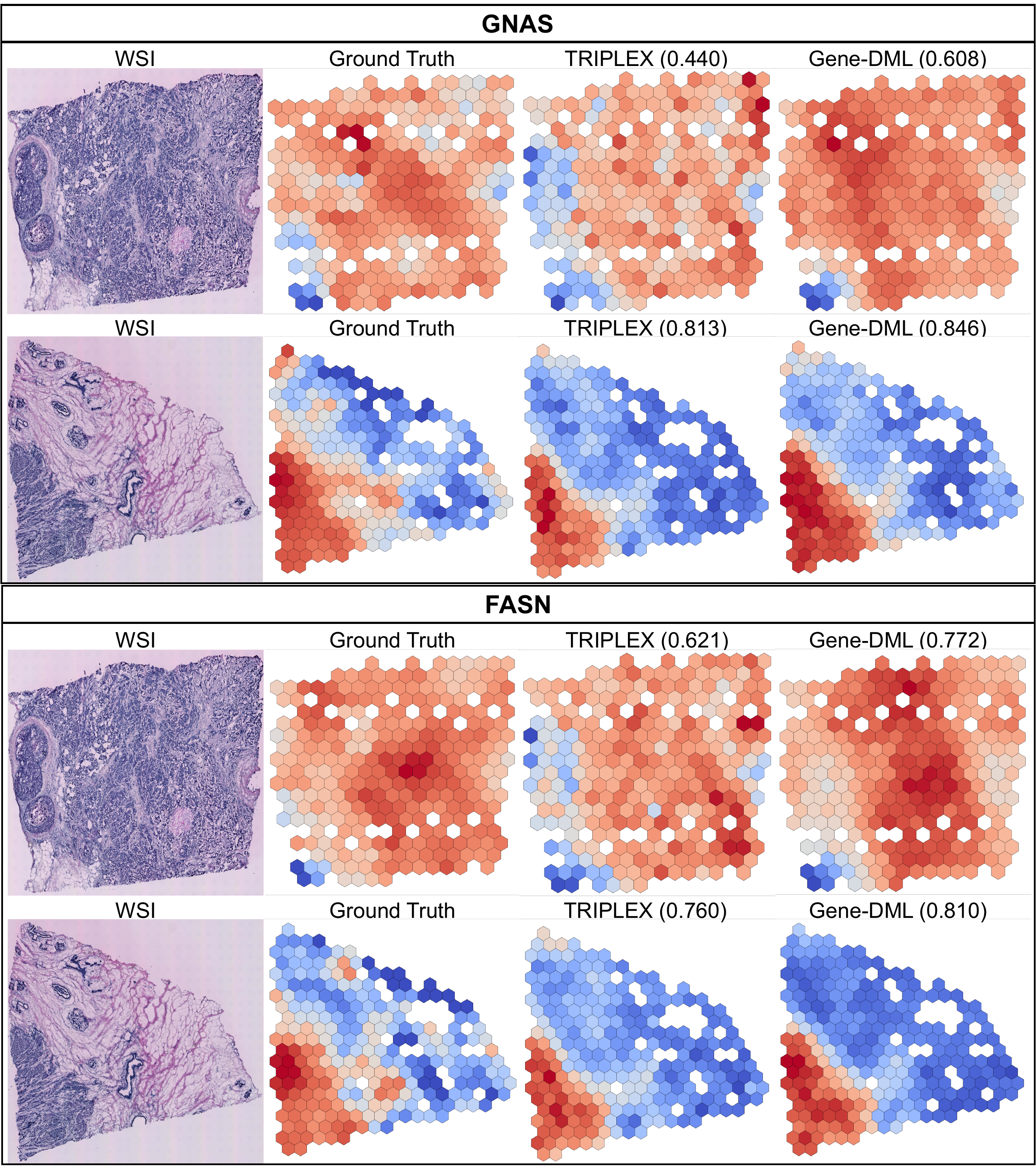}
    \caption{Visualization of cancer biomarker gene expression level prediction. All predicted values of gene expression are normalized to the range $[0,1]$. The values in parentheses are the PCC between the ground truth of gene expression and the prediction on genes GNAS and FASN.}
    \label{fig3}
    \vspace{-0.2cm}
\end{figure}

\subsection{Generalization Performance of Gene-DML}

\cref{tab2_general_main} presents the generalization performance of Gene-DML compared to recent existing works across $3$ unseen Visium ST datasets.
The results are produced by using the pre-trained Gene-DML model on the breast cancer dataset (HER2ST).
Our method achieves the highest Pearson Correlation Coefficient (PCC) on both PCC(A) and PCC(H) genes across all datasets, highlighting its superior ability to capture biologically meaningful patterns that generalize beyond training distributions. 
Notably, Gene-DML achieves a new state-of-the-art PCC(A) of $0.160$, $0.192$, $0.155$ and PCC(H) of $0.415$, $0.411$, $0.393$ on breast Visium-1, -2, and -3, respectively, surpassing all prior approaches.
While our MSE and MAE scores are not always the best, this can be attributed to Gene-DML’s emphasis on semantic alignment and structure-preserving embeddings rather than purely minimizing per-spot regression loss. The slight trade-off in MSE/MAE is acceptable, as PCC offers a more interpretable measure of functional correlation, especially for HPGs critical in downstream biological interpretation.
The result of the additional $2$ skin Visium data (NCBI463 and NCBI464) is included in the supplementary.

\subsection{Visualization of Cancer Biomarker Genes}
GNAS~\cite{cancers14225480} and FASN~\cite{jiang2021met} are two of the highly predictive genes across all datasets we utilized, and well-established cancer biomarker genes.
\cref{fig3} presents a qualitative comparison of gene expression prediction with indicating PCC values for GNAS and FASN across two representative tissue WSIs from various patients, with the original WSI, gene expression ground truth, TRIPLEX prediction, and Gene-DML prediction. 
Specifically, each hexagonal tile corresponds to a spatial transcriptomics (ST) spot, with its color indicating the predicted gene expression level, normalized to the range $[0, 1]$.
Gene-DML consistently outperforms the model TRIPLEX with the second-best quantitative results in~\cref{tab1_baseline} in accurately recovering the spatial expression patterns, as evidenced by the visibly sharper regional expression and higher PCC values (\eg, $0.608$ vs. $0.440$ and $0.846$ vs. $0.813$ for GNAS; $0.772$ vs. $0.621$ and $0.810$ vs. $0.760$ for FASN) for both relatively hard (sample \#1) and easy (sample \#2) cases. 
The robust prediction of GNAS and FASN not only reflects the strength of features learned by the dual multi-scale discrimination but also is widely implicated in tumorigenesis and cancer progression, further underscoring the biological relevance and clinical utility of the Gene-DML framework for spatial biomarker discovery.
Visualizations of additional samples are provided in the supplementary material.

\begin{table}[t]
\centering
\resizebox{\linewidth}{!}{
\setlength{\tabcolsep}{0.4mm}{
\begin{tabular}{cccc|cccc}
\toprule
\multicolumn{1}{c}{\makecell[c]{Multi-scale\\ I-L Dis.}} & \multicolumn{1}{c}{\makecell[c]{Global\\Rep.}} & \multicolumn{1}{c}{\makecell[c]{Cross-level\\ I-G Dis.}} & \multicolumn{1}{c|}{\makecell[c]{Feature\\ Grouping}} & MSE$\downarrow$ & MAE$\downarrow$ & PCC(A)$\uparrow$ & PCC(H)$\uparrow$ \\
\midrule
\ding{55}&\ding{55} & \ding{55}&\ding{55}& 0.297 & 0.430 & 0.269 & 0.396\\
\ding{51}&\ding{51}&\ding{55} & \ding{55} &0.282 & 0.414 & 0.308 & 0.449\\
\ding{55}&\ding{55}&\ding{51}&\ding{51} & 0.280 & 0.413 & 0.320 & 0.478\\
\ding{51}&\ding{55}&\ding{51}&\ding{51} & 0.247 & 0.394 &0.421 & 0.478\\
\ding{51}&\ding{51}&\ding{51} & \ding{55} & 0.253 & 0.399 & 0.398 & 0.456\\
\rowcolor{gray!25} \ding{51}&\ding{51} &\ding{51}&\ding{51} & \textbf{0.237} & \textbf{0.350} & \textbf{0.433} & \textbf{0.520} \\
\bottomrule
\end{tabular}
}}
\caption{Ablation study on framework components on the dataset skinST. Abbreviations in this table include instance-level discrimination (I-L Dis.), global-scale representation (Global Rep.), and instance-group discrimination (I-G Dis.).}
\label{tab3_abl}
\end{table}

\begin{table}[t]
\centering
\resizebox{\linewidth}{!}{
\setlength{\tabcolsep}{4.3mm}{
\begin{tabular}{c|cccc}
\toprule
$k$ & MSE$\downarrow$ & MAE$\downarrow$ & PCC(A)$\uparrow$ & PCC(H)$\uparrow$ \\
\midrule
15 & 0.252 & 0.363 & 0.426 & 0.516\\
20 & 0.250 & 0.363 & 0.433 & 0.503\\
\rowcolor{gray!25}25 & \textbf{0.237} & \textbf{0.350} & \textbf{0.433} & \textbf{0.520} \\
30 & 0.240 & 0.352 & 0.429 & 0.513\\
\bottomrule
\end{tabular}
}}
\caption{Ablation study on the number of $k$-groups in the feature grouping process on the dataset skinST.}
\label{tab4_abl}
\end{table}

\subsection{Ablation Study}

We performed extensive experiments to assess the effectiveness of each framework module with its key component, values of hyperparameters $k$ (feature group size), $\lambda$ (the weight of instance-group alignment loss), and $\tau_{I-G}$ (the temperature of instance-group similarity calculation).

We report the ablation results on each key component of Gene-DML, as illustrated in~\cref{tab3_abl}. 
This underscores the effectiveness of the \textbf{multi-scale instance-level discrimination} module and the \textbf{cross-level instance-group discrimination} module, together with their corresponding \textit{key components}: the \textbf{global-scale representation} and \textbf{feature grouping}.
Incorporating the multi-scale instance-level discrimination improves performance $0.015$ on MSE, $+0.039$ on PCC(A), and $+0.053$ on PCC(H). 
Similarly, adding cross-level instance-group discrimination yields a gain $0.017$ on MSE, $0.051$ on PCC(A), and $0.082$ on PCC(H) from the plain bimodal framework.
When both modules are integrated, Gene-DML achieves the best performance across all metrics, showing a cumulative improvement of $0.060$ MSE, $+0.164$ PCC(A), and $+0.124$ PCC(H) over the baseline. 
Besides, comparative experiments with and without the global-scale representation and feature grouping further highlight their essential contribution to the effectiveness of the corresponding modules.
These results validate the complementary benefits of a dual-pathway multi-level discrimination strategy.

\begin{table}[t]
\centering
\resizebox{\linewidth}{!}{
\setlength{\tabcolsep}{4mm}{
\begin{tabular}{c|cccc}
\toprule
 $\lambda$ &  MSE$\downarrow$ & MAE$\downarrow$ & PCC(A)$\uparrow$ & PCC(H)$\uparrow$ \\
\midrule
0.3 & 0.263 & 0.380 & 0.391 & 0.430\\
0.5 & 0.259 & 0.373 & 0.402 & 0.493\\
\rowcolor{gray!25}0.8 & \textbf{0.237} & \textbf{0.350} & \textbf{0.433} & \textbf{0.520} \\
1 & 0.241 & 0.365 & 0.426 & 0.510\\
\bottomrule
\end{tabular}
}}
\caption{Ablation study on the weight of instance-group loss $\lambda$ on the dataset skinST. $\lambda=1$ indicates that the weight of instance-group loss is not applied in the experimental setting.}
\label{tab5_abl}
\end{table}

\begin{table}[t]
\centering
\resizebox{\linewidth}{!}{
\setlength{\tabcolsep}{4mm}{
\begin{tabular}{c|cccc}
\toprule
$\tau_{I\text{-}G}$ & MSE$\downarrow$ & MAE$\downarrow$ & PCC(A)$\uparrow$ & PCC(H)$\uparrow$ \\
\midrule
0.05 & 0.269 & 0.380 & 0.420 & 0.513\\
\rowcolor{gray!25}0.07 & \textbf{0.237} & \textbf{0.350} & \textbf{0.433} & \textbf{0.520} \\
0.1 & 0.273 & 0.381 & 0.412 & 0.490\\
\bottomrule
\end{tabular}
}}
\caption{Ablation study on the temperature of instance-group similarity calculation $\tau_{I\text{-}G}$ on the dataset skinST.}
\label{tab6_abl}
\end{table}

\cref{tab4_abl} demonstrates the impact of varying the number of groups $k$ in the feature grouping (\cref{sec:feature_grouping}). The model performance is relatively robust across different values but achieves the best results when $k = 25$, with the lowest MSE and MAE, and the highest PCC(A) and PCC(H). 
These results suggest that an appropriately chosen group size provides a good balance between semantic expressiveness and discrimination, facilitating more effective instance-group alignment.
\cref{tab5_abl} presents an ablation study on the impact of the weighting coefficient $\lambda$ for the instance-group contrastive loss. The results show that incorporating instance-group discrimination consistently improves performance across all metrics. 
Specifically, when $\lambda = 0.8$, the best overall results are obtained, achieving the lowest MSE and MAE, along with the highest PCC scores. This indicates that moderately weighting the instance-group loss effectively enhances cross-modal instance-group alignment without overwhelming the training objective.
\cref{tab6_abl} shows the effect of the temperature parameter $\tau_{I\text{-}G}$ used in the instance-group similarity calculation. 
$\tau_{I\text{-}G}$ controls the sharpness of the softmax distribution over similarity scores. The results show that setting $\tau_{I\text{-}G} = 0.07$ achieves optimal performance across all evaluation metrics.
The ablation of HER2ST and STNet is included in the supplementary.

\section{Conclusion}

In this work, we introduced Gene-DML, a novel framework that advances gene expression prediction from histopathology images by modeling dual-pathway multi-level discrimination. 
By jointly integrating multi-scale instance-level and cross-level instance-group discriminations, Gene-DML effectively captures both fine-grained and high-level semantic correspondences across visual and transcriptomic modalities. 
Validated through extensive experiments, our dual-pathway discrimination design facilitates robust and generalizable cross-modal representation learning in spatial transcriptomic data, enabling the model to better reflect the biological and spatial complexities inherent in tissue samples.

\newpage
{
    \small
    \bibliographystyle{ieeenat_fullname}
    \bibliography{main}
}
\clearpage
\maketitlesupplementary
\appendix

\noindent This supplementary material complements the main manuscript by providing detailed information and additional support. It is structured as follows:

\paragraph{\cref{supp:mae}: Additional Cross-validation Performance}
- Provides the MAE results on three ST datasets to further evaluate the cross-validation performance of Gene-DML.

\paragraph{\cref{supp:general}: Additional Generalization Performance}
- Provides the generalization performance of Gene-DML compared to TRIPLEX across $2$ additional unseen skin cancer Visium ST data.

\paragraph{\cref{supp:abl}: Additional Ablation Study}
- Provides the effectiveness of each key component in the model Gene-DML and hyperparameter configurations of $k$, $\lambda$, and $\tau_{I\text{-}G}$ on datasets HER2ST and STNet.

\paragraph{\cref{supp:vis}: Additional Visualization}
- Provides additional sample visualizations of gene expression prediction of cancer biomarker genes across three ST datasets.

\paragraph{\cref{supp:gene}: Additional Implementation Details}
- Provides specific $250$ genes selected for each dataset to evaluate PCC(A) metric, the average of the Pearson Correlation Coefficient of $250$ selected genes, of the Gene-DML.

\section{Additional Cross-validation Performance}

\label{supp:mae}

\cref{tab7_mae} demonstrates a cross-validation comparison of mean absolute error (MAE) across three spatial transcriptomics datasets, HER2ST, STNet, and skinST, benchmarking our proposed method, Gene-DML, against several state-of-the-art baseline methods. 
Gene-DML consistently achieves the best performance across all datasets. 
Specifically, on HER2ST, Gene-DML attains an MAE of $0.350$, outperforming the second-best method, TRIPLEX ($0.362$), with $0.012$ improvement. 
On STNet, Gene-DML achieves an MAE of $0.327$, improving upon Hist2ST ($0.333$) by $0.006$. On skinST, Gene-DML reaches an MAE of $0.350$, outperforming TRIPLEX ($0.404$) with a substantial $0.054$ improvement. 
These consistent gains across diverse datasets highlight the effectiveness of the dual-dimensional multi-level discrimination framework in capturing semantically aligned multi-level cross-modal representations.

\begin{table}[ht]
\centering
\begin{adjustbox}{width=\linewidth}
\label{t4}
\footnotesize
\begin{tabular}{c|c|c|c}
\toprule
\multirow{2}[2]{*}{Model} & \multicolumn{1}{c|}{HER2ST} & \multicolumn{1}{c|}{STNet} & \multicolumn{1}{c}{skinST} \\
\cmidrule(lr){2-4}
& MAE$\downarrow$ & MAE$\downarrow$ & MAE$\downarrow$ \\
\midrule
ST-Net \cite{he_2020} & $0.389\pm0.03$ & $0.349\pm0.02$ & $0.428\pm0.05$ \\
HistoGene \cite{pang_2021} & $0.428\pm0.07$ & $0.335\pm0.04$ & $0.415\pm0.07$ \\
Hist2ST \cite{zeng_2022} & $0.413\pm0.07$ & $\underline{0.333}\pm0.02$ & $0.924\pm0.29$ \\
EGN \cite{Yang_2023} & $0.377\pm0.04$ & $0.337\pm0.02$ & $0.418\pm0.06$ \\
BLEEP \cite{xie_2024} & $0.401\pm0.03$ & $0.369\pm0.02$ & $0.430\pm0.04$ \\
TRIPLEX \cite{chung_2024} & $\underline{0.362}\pm0.05$ & $0.343\pm0.02$ & $\underline{0.404}\pm0.07$ \\
M2OST \cite{wang2025m2ost} & $0.421\pm0.02$ & $0.362\pm0.01$ & $0.418\pm0.03$ \\
\rowcolor{gray!25} Gene-DML& $\textbf{0.350}\pm0.04$ & $\textbf{0.327}\pm0.02$ & $\textbf{0.350}\pm0.05$ \\
\bottomrule
\end{tabular}
\end{adjustbox}
\caption{Additional comparison of cross-validation experiments with baselines on metric MAE. The \textbf{best results} are in bold; the \underline{second-best results} are underlined. 
}
\label{tab7_mae}
\end{table}

\section{Additional Generalization Performance}
\label{supp:general}

\begin{table}[ht]
\setlength{\tabcolsep}{0.3mm}{
\resizebox{\linewidth}{!}{%
\begin{tabular}{c|cccc|cccc}
\toprule[1pt]
\multirow{2}[2]{*}{Model} & \multicolumn{4}{c|}{NCBI463} & \multicolumn{4}{c}{NCBI464} \\
\cmidrule(lr){2-5}
\cmidrule(lr){6-9}
& MSE$\downarrow$ & MAE$\downarrow$ & PCC(A)$\uparrow$ & PCC(H)$\uparrow$ & MSE$\downarrow$ & MAE$\downarrow$ & PCC(A)$\uparrow$ & PCC(H)$\uparrow$ \\
\midrule
TRIPLEX& 0.342 & 0.478 & 0.103 & 0.360 &
0.416 & 0.538 & 0.091 & 0.261 \\
\textbf{Gene-DML (ours)}& \textbf{0.253} & \textbf{0.413} & \textbf{0.109} & \textbf{0.377} &
\textbf{0.292} & \textbf{0.445} & \textbf{0.140} & \textbf{0.441} \\
\bottomrule[1pt]
\end{tabular}}
\caption{Generalization performance on NCBI463 and NCBI464.}
\label{tab2_general}
}
\end{table}

We further extend the evaluation by adding two skin cancer Visium datasets (NCBI463 and NCBI464)~\cite{Jaume_2025} to verify the skinST pre-trained models.
\cref{tab2_general} further underscores Gene-DML's outstanding generalization performance, achieving state-of-the-art results across all evaluation metrics MSE, MAE, PCC(A), and PCC(H).

\section{Additional Ablation Study}
\label{supp:abl}

To evaluate the dual-pathway multi-level discrimination framework, Gene-DML, we perform an ablation study by adding each discrimination pathway sequentially together with their corresponding key components: the global-scale representation and feature grouping, and report the results of MSE, MAE, PCC(A), and PCC(H) metrics. \cref{tab1_abl_her2st} and \cref{tab1_abl_stnet} demonstrate the effectiveness of each component in Gene-DML on datasets HER2ST and STNet.
They show that, by removing both pathways, a plain bimodal framework yields the weakest performance.
By incorporating the multi-scale instance-level discrimination pathway and cross-level instance-group discrimination pathway, respectively, a significant improvement is achieved.

For the hyperparameter configuration of $k$ (number of groups), $\lambda$ (the weight of instance-group cross-level loss), and $\tau_{I\text{-}G}$ (the temperature of instance-group similarity calculation), \cref{tab2_abl_her2st}, \cref{tab3_abl_her2st}, and \cref{tab4_abl_her2st} respectively present the result of the dataset HER2ST. As for the STNet dataset, please refer to \cref{tab2_abl_stnet}, \cref{tab3_abl_stnet}, and \cref{tab4_abl_stnet}.

Overall, all these results show that while $k = 18$ and $\lambda = 0.8$ with $\tau_{I\text{-}G} = 0.07$, Gene-DML could achieve optimal performance across all evaluation metrics, thus setting it as the default configuration for HER2ST dataset; besides, $k = 25$ and $\lambda = 0.8$ with $\tau_{I\text{-}G} = 0.07$ are set as the default configuration for STNet dataset.

\begin{table}[ht]
\centering
\resizebox{\linewidth}{!}{
\setlength{\tabcolsep}{0.5mm}{
\begin{tabular}{cccc|cccc}
\toprule
\multicolumn{1}{c}{\makecell[c]{Multi-scale\\ I-L Dis.}} &\multicolumn{1}{c}{\makecell[c]{Global\\Rep.}} & \multicolumn{1}{c}{\makecell[c]{Cross-level\\ I-G Dis.}} & \multicolumn{1}{c|}{\makecell[c]{Feature\\Grouping}} & MSE$\downarrow$ & MAE$\downarrow$ & PCC(A)$\uparrow$ & PCC(H)$\uparrow$ \\
\midrule
\ding{55}&\ding{55} & \ding{55}&\ding{55}& 0.340 & 0.445 & 0.231 & 0.392\\
\ding{51}&\ding{51}&\ding{55} & \ding{55} & 0.298 & 0.412 & 0.301 & 0.463\\
\ding{55}&\ding{55}&\ding{51}&\ding{51}&0.270 & 0.415 & 0.302 & 0.482\\
\ding{51}&\ding{55}&\ding{51}&\ding{51} & 0.221 & 0.364 & 0.282& 0.416 \\
\ding{51}&\ding{51}&\ding{51} & \ding{55} & 0.230 &0.360&0.303 & 0.400\\
\rowcolor{gray!25} \ding{51}&\ding{51} &\ding{51}&\ding{51} & \textbf{0.210} & \textbf{0.350}&\textbf{0.331} & \textbf{0.541} \\
\bottomrule
\end{tabular}
}}
\caption{Ablation study on components of the dataset HER2ST. Abbreviations in this table include instance-level discrimination (I-L Dis.), global-scale representation (Global Rep.) and instance-group discrimination (I-G Dis.).}
\label{tab1_abl_her2st}
\end{table}

\begin{table}[ht]
\centering
\resizebox{\linewidth}{!}{
\setlength{\tabcolsep}{4.3mm}{
\begin{tabular}{c|cccc}
\toprule
$k$ & MSE$\downarrow$ & MAE$\downarrow$ & PCC(A)$\uparrow$ & PCC(H)$\uparrow$ \\
\midrule
15 & 0.242 & 0.371 & 0.299 & 0.510\\
16 & 0.235 & 0.370 & 0.323 & 0.536\\
\rowcolor{gray!25} 18 & \textbf{0.210} & \textbf{0.350}&\textbf{0.331} & \textbf{0.541} \\
20 & 0.250 & 0.397 & 0.320 & 0.514\\
\bottomrule
\end{tabular}
}}
\caption{Ablation study on the number of $k$-groups in the feature grouping process on the dataset HER2ST.}
\label{tab2_abl_her2st}
\end{table}

\begin{table}[ht]
\centering
\resizebox{\linewidth}{!}{
\setlength{\tabcolsep}{4mm}{
\begin{tabular}{c|cccc}
\toprule
 $\lambda$ &  MSE$\downarrow$ & MAE$\downarrow$ & PCC(A)$\uparrow$ & PCC(H)$\uparrow$ \\
\midrule
0.3 & 0.231 & 0.369 & 0.326 & 0.525\\
0.5 & 0.213 & 0.359 & 0.331 & 0.532\\
\rowcolor{gray!25} 0.8 & \textbf{0.210} & \textbf{0.350}&\textbf{0.331} & \textbf{0.541} \\
1 & 0.215 & 0.360 & 0.322 & 0.537\\
\bottomrule
\end{tabular}
}}
\caption{Ablation study on the weight of instance-group loss $\lambda$ on the dataset HER2ST. $\lambda=1$ indicates that the weight of instance-group loss is not applied in the experimental setting.}
\label{tab3_abl_her2st}
\end{table}

\begin{table}[ht]
\centering
\resizebox{\linewidth}{!}{
\setlength{\tabcolsep}{4mm}{
\begin{tabular}{c|cccc}
\toprule
$\tau_{I\text{-}G}$ & MSE$\downarrow$ & MAE$\downarrow$ & PCC(A)$\uparrow$ & PCC(H)$\uparrow$ \\
\midrule
0.05 & 0.220 & 0.355 & 0.329 & 0.532\\
\rowcolor{gray!25} 0.07 & \textbf{0.210} & \textbf{0.350}&\textbf{0.331} & \textbf{0.541} \\
0.1 & 0.238 & 0.371 & 0.320 & 0.534\\
\bottomrule
\end{tabular}
}}
\caption{Ablation study on the temperature of instance-group similarity calculation $\tau_{I\text{-}G}$ on the dataset HER2ST.}
\label{tab4_abl_her2st}
\end{table}

\begin{table}[ht]
\centering
\resizebox{\linewidth}{!}{
\setlength{\tabcolsep}{0.5mm}{
\begin{tabular}{cccc|cccc}
\toprule
\multicolumn{1}{c}{\makecell[c]{Multi-scale\\ I-L Dis.}} &\multicolumn{1}{c}{\makecell[c]{Global\\Rep.}} & \multicolumn{1}{c}{\makecell[c]{Cross-level\\ I-G Dis.}} & \multicolumn{1}{c|}{\makecell[c]{Feature\\Grouping}} & MSE$\downarrow$ & MAE$\downarrow$ & PCC(A)$\uparrow$ & PCC(H)$\uparrow$ \\
\midrule
\ding{55}&\ding{55} & \ding{55}&\ding{55}& 0.269 & 0.440 & 0.160 & 0.233\\
\ding{51}&\ding{51}&\ding{55} & \ding{55} & 0.246 & 0.400 & 0.211 & 0.308\\
\ding{55}&\ding{55}&\ding{51}&\ding{51} & 0.232 & 0.379 & 0.198 & 0.301\\
\ding{51}&\ding{55}&\ding{51}&\ding{51} & 0.184 & 0.339 &0.227 &0.361\\
\ding{51}&\ding{51}&\ding{51} & \ding{55} & 0.188 & 0.338 & 0.216 & 0.309 \\
\rowcolor{gray!25} \ding{51}&\ding{51} &\ding{51}&\ding{51} & \textbf{0.179} & \textbf{0.327} & \textbf{0.237}& \textbf{0.384} \\
\bottomrule
\end{tabular}
}}
\caption{Ablation study on components of the dataset STNet. Abbreviations in this table include instance-level discrimination (I-L Dis.), global-scale representation (Global Rep.) and instance-group discrimination (I-G Dis.).}
\label{tab1_abl_stnet}
\end{table}

\begin{table}[ht]
\centering
\resizebox{\linewidth}{!}{
\setlength{\tabcolsep}{4.3mm}{
\begin{tabular}{c|cccc}
\toprule
$k$ & MSE$\downarrow$ & MAE$\downarrow$ & PCC(A)$\uparrow$ & PCC(H)$\uparrow$ \\
\midrule
80 & 0.185 & 0.331 & 0.236 & 0.376\\
\rowcolor{gray!25} 90 & \textbf{0.179} & \textbf{0.327} & \textbf{0.237}& \textbf{0.384} \\
100 & 0.188 & 0.332 & 0.230 & 0.368\\
\bottomrule
\end{tabular}
}}
\caption{Ablation study on the number of $k$-groups in the feature grouping process on the dataset STNet.}
\label{tab2_abl_stnet}
\end{table}

\begin{table}[ht]
\centering
\resizebox{\linewidth}{!}{
\setlength{\tabcolsep}{4mm}{
\begin{tabular}{c|cccc}
\toprule
 $\lambda$ &  MSE$\downarrow$ & MAE$\downarrow$ & PCC(A)$\uparrow$ & PCC(H)$\uparrow$ \\
\midrule
0.5 & 0.205 & 0.350 & 0.200 & 0.349\\
\rowcolor{gray!25} 0.8 & \textbf{0.179} & \textbf{0.327} & \textbf{0.237}& \textbf{0.384} \\
1 & 0.202 & 0.345 & 0.228 & 0.361\\
\bottomrule
\end{tabular}
}}
\caption{Ablation study on the weight of instance-group loss $\lambda$ on the dataset STNet. $\lambda=1$ indicates that the weight of instance-group loss is not applied in the experimental setting.}
\label{tab3_abl_stnet}
\end{table}

\begin{table}[ht]
\centering
\resizebox{\linewidth}{!}{
\setlength{\tabcolsep}{4mm}{
\begin{tabular}{c|cccc}
\toprule
$\tau_{I\text{-}G}$ & MSE$\downarrow$ & MAE$\downarrow$ & PCC(A)$\uparrow$ & PCC(H)$\uparrow$ \\
\midrule
0.05 & 0.201 & 0.365 & 0.222 & 0.375\\
\rowcolor{gray!25} 0.07 & \textbf{0.179} & \textbf{0.327} & \textbf{0.237}& \textbf{0.384} \\
0.1 & 0.186 & 0.331 & 0.235 & 0.382\\
\bottomrule
\end{tabular}
}}
\caption{Ablation study on the temperature of instance-group similarity calculation $\tau_{I\text{-}G}$ on the dataset STNet.}
\label{tab4_abl_stnet}
\end{table}

\section{Additional Visualizations}
\label{supp:vis}

\begin{figure*}[ht]
    \centering
    \includegraphics[width=0.58\textwidth]{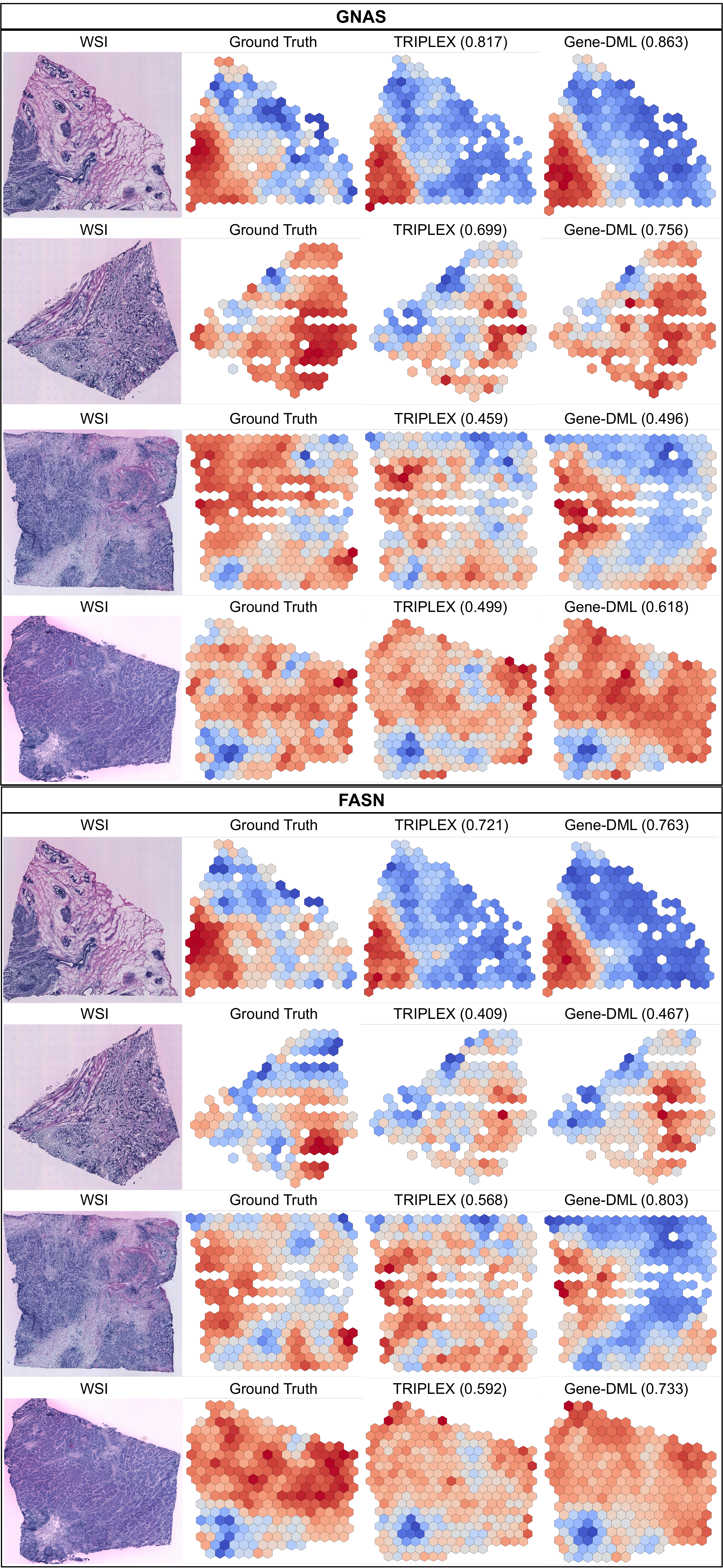}
    \caption{Visualization of cancer biomarker genes GNAS and FASN prediction on the dataset HER2ST.}
    \label{her2st_vis}
\end{figure*}

\begin{figure*}[ht]
    \centering
    \includegraphics[width=0.58\textwidth]{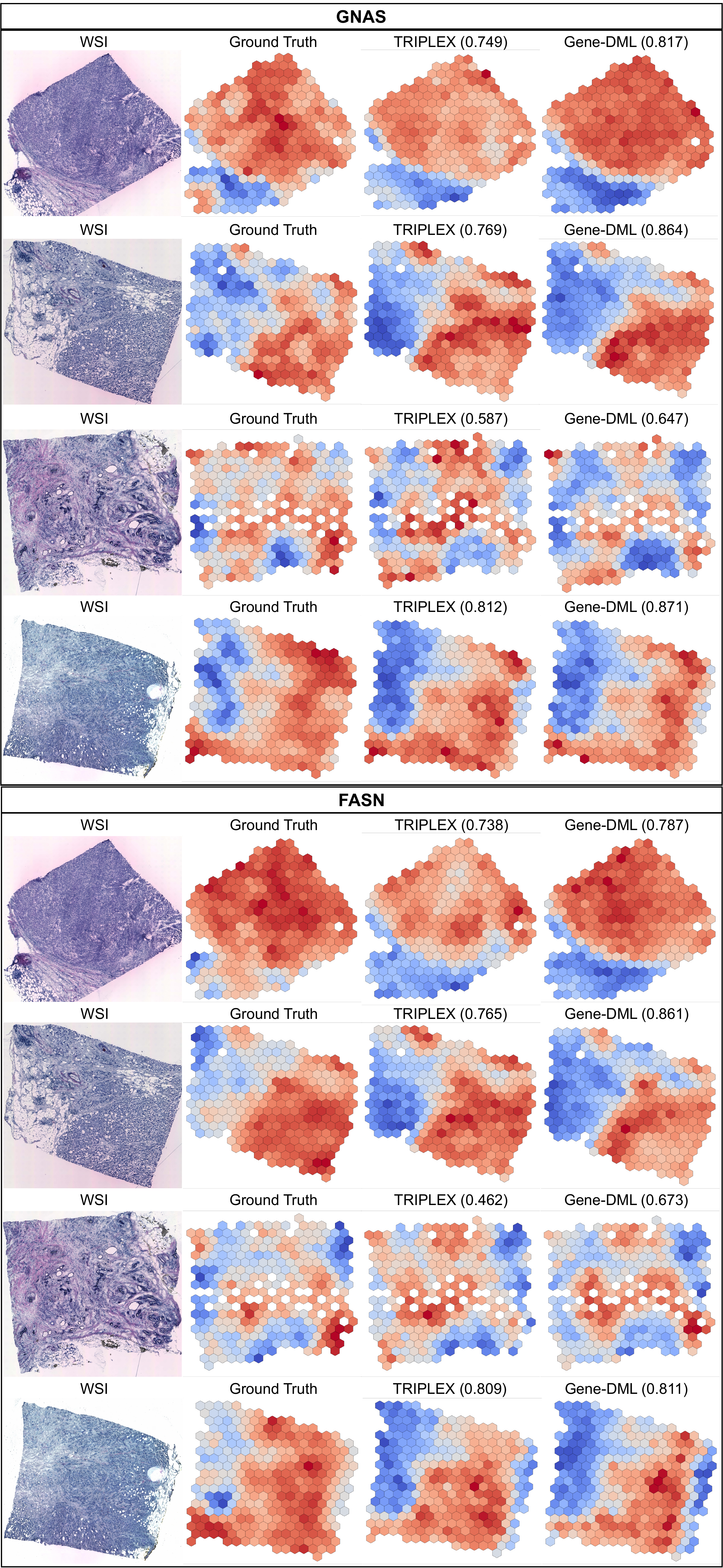}
    \caption{Visualization of cancer biomarker genes GNAS and FASN prediction on the dataset STNet.}
    \label{stnet_vis}
\end{figure*}

In this section, we provide additional visualizations of the gene expression prediction results, which are normalized into the range $[0,1]$, of the breast cancer biomarker genes GNAS and FASN.
\cref{her2st_vis} demonstrates the spatial gene expression prediction for GNAS and FASN on the HER2ST dataset.
Each row displays a WSI sample alongside the ground-truth expression map and predictions from TRIPLEX (the method achieved the second-best result evaluated in Tab. 1 of Sec. 4.2.) and the proposed Gene-DML.
Across both genes and multiple WSI samples, Gene-DML consistently produces spatial expression patterns that are visually closer to the ground truth and quantitatively higher in performance, as reflected by the reported Pearson correlation coefficients in parentheses.
The same conclusion holds for \cref{stnet_vis}, which shows the spatial gene expression prediction results on the dataset STNet, further validating the strong gene expression prediction capacity of Gene-DML.

\section{Additional Implementation Details}
\label{supp:gene}

\begin{figure*}[ht]
    \centering
    \includegraphics[width=\textwidth]{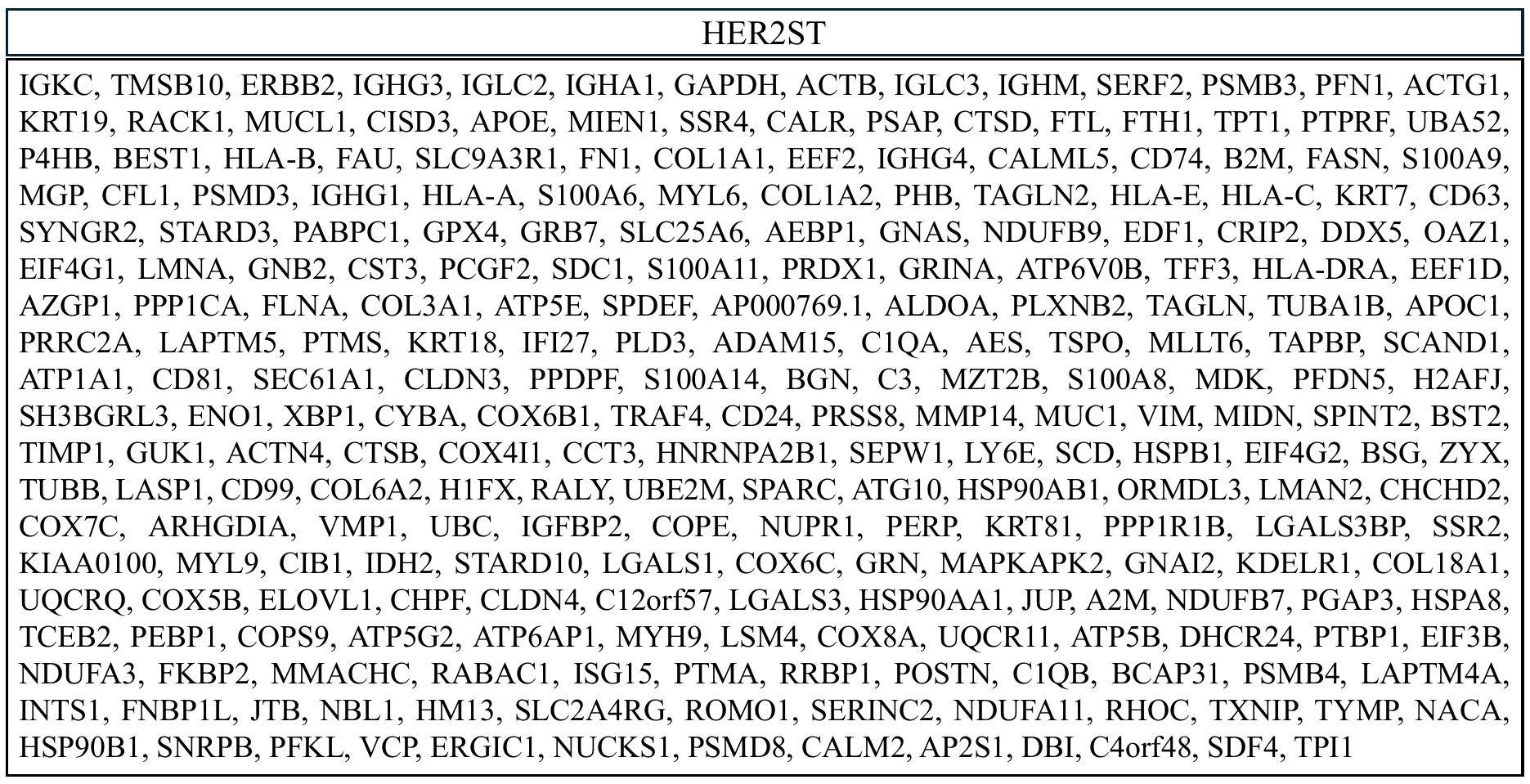}
    \caption{$250$ Genes for the dataset HER2ST.}
    \label{her2st_gene}
\end{figure*}

\begin{figure*}[ht]
    \centering
    \includegraphics[width=\textwidth]{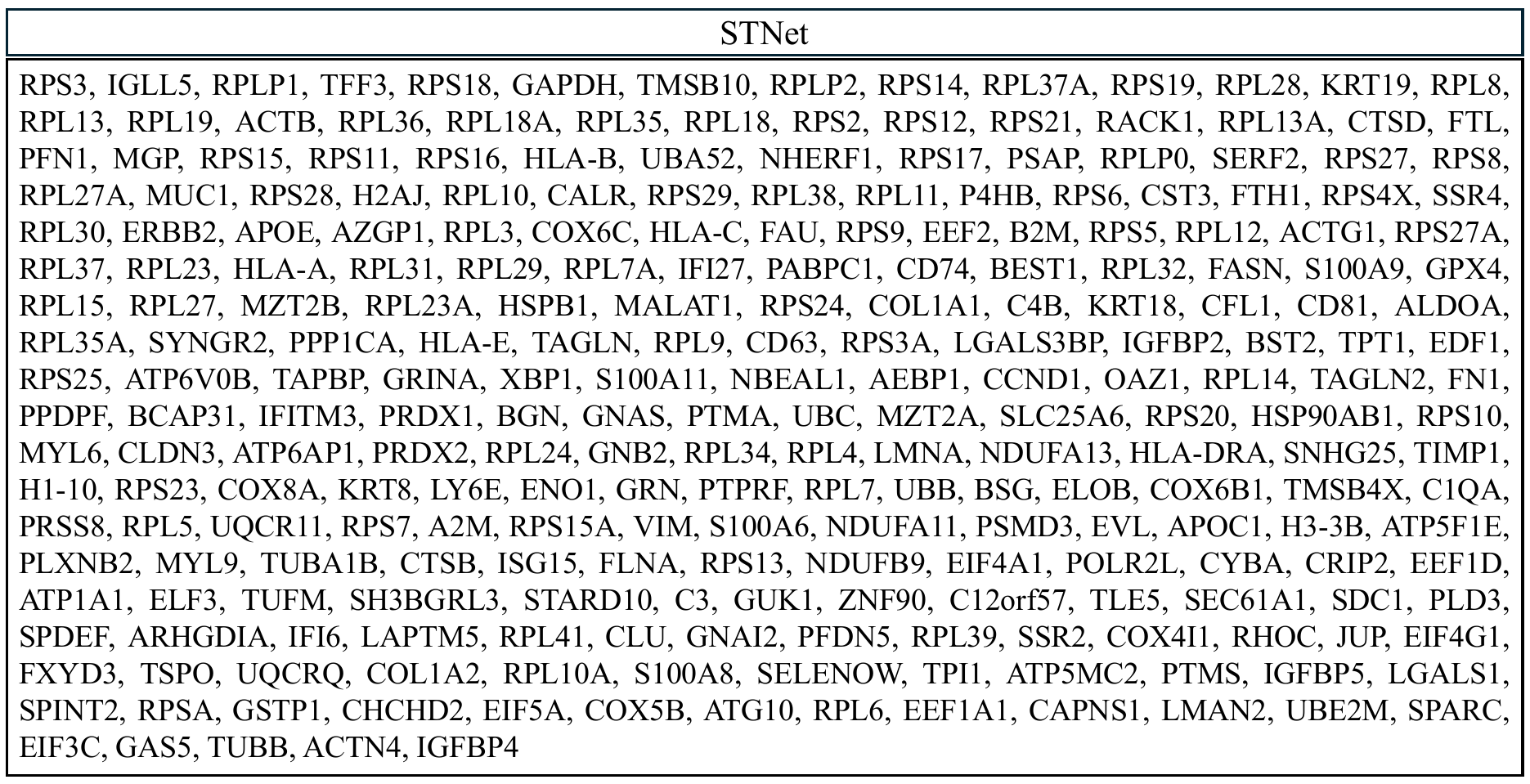}
    \caption{$250$ Genes for the dataset STNet.}
    \label{stnet_gene}
\end{figure*}

\begin{figure*}[ht]
    \centering
    \includegraphics[width=\textwidth]{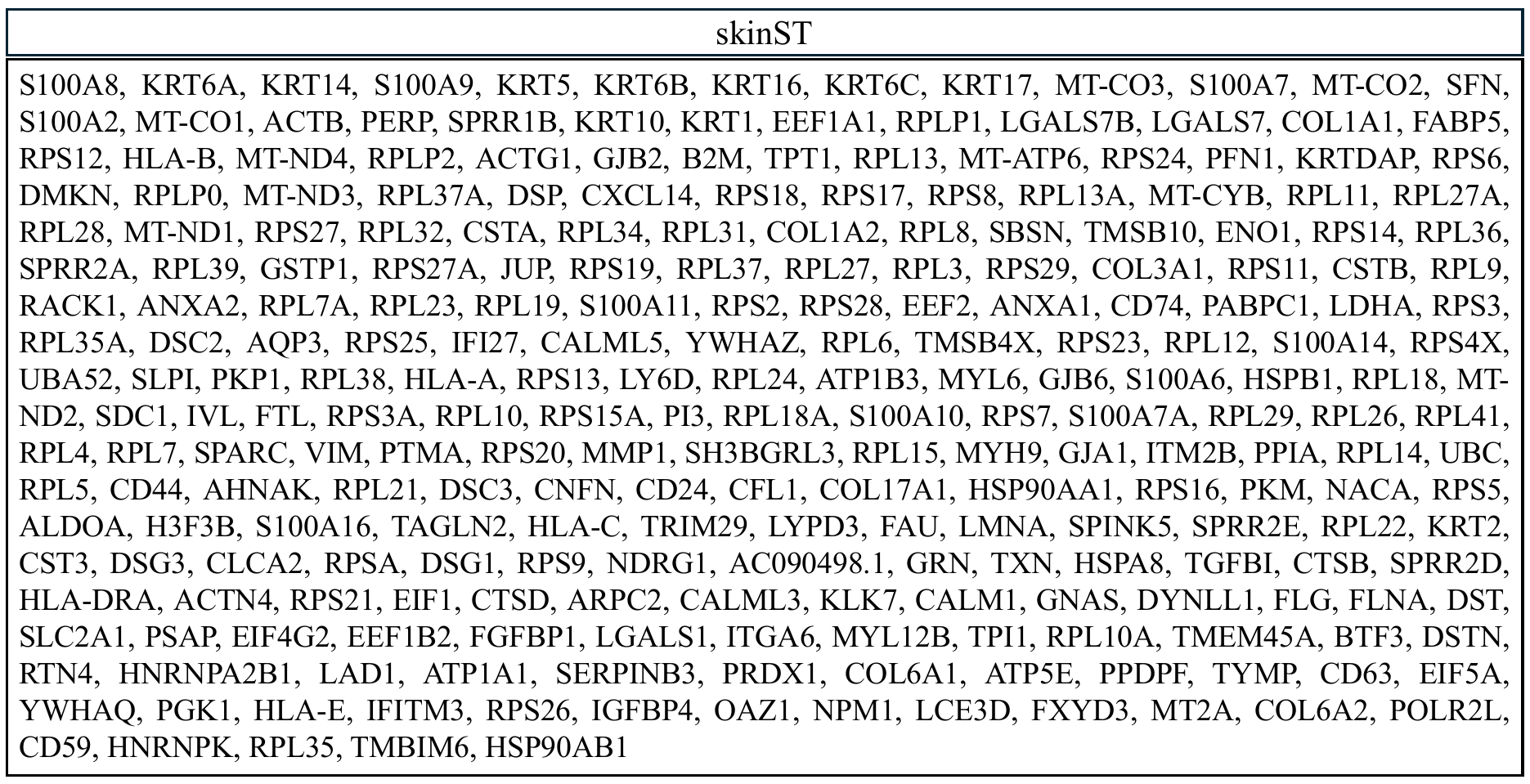}
    \caption{$250$ Genes for the dataset skinST.}
    \label{skin_gene}
\end{figure*}

To ensure fairness, following TRIPLEX~\cite{chung_2024}, we utilize the various $250$ selected genes for each dataset, \ie, HER2ST, STNet, and skinST, that are related to morphology in histopathology to evaluate the performance of gene expression prediction.
For the results of the MSE, MAE, and PCC(A) metrics in the cross-validation experiments, the same $250$ genes were utilized for both the training and testing processes.
\cref{her2st_gene}, \cref{stnet_gene}, and \cref{skin_gene} demonstrate the $250$ genes exploited to predict gene expression profiles on datasets HER2ST, STNet, and skinST, respectively. 

\end{document}